\documentclass{article}

\usepackage[nonatbib,preprint]{neurips_2022}
\usepackage[utf8]{inputenc} 
\usepackage[T1]{fontenc}    
\usepackage{url}            
\usepackage{booktabs}       
\usepackage{amsfonts}       
\usepackage{nicefrac}       
\usepackage{microtype}      
\usepackage[dvipsnames,table,xcdraw]{xcolor}
\usepackage[numbers,sort]{natbib}
\usepackage{multirow}
\usepackage{nicematrix}
\usepackage{graphicx}
\usepackage{bbding}
\usepackage{floatrow}
\usepackage{subcaption}
\usepackage[font=small]{caption}
\usepackage{pifont}
\usepackage{wrapfig}
\usepackage{hyperref}
\usepackage{tabularx}

\usepackage{arydshln}

\newfloatcommand{capbtabbox}{table}[][\FBwidth]
\newcommand{\tablestyle}[2]{\setlength{\tabcolsep}{#1}\renewcommand{\arraystretch}{#2}\centering\footnotesize}
\newlength\savewidth\newcommand\shline{\noalign{\global\savewidth\arrayrulewidth
		\global\arrayrulewidth 1pt}\hline\noalign{\global\arrayrulewidth\savewidth}}

\hypersetup{
    colorlinks=true,
    linkcolor=red,
    citecolor=RoyalBlue,
    filecolor=magenta,      
    urlcolor=cyan,
}

\title{CMAE-V: Contrastive Masked Autoencoders for \\ Video Action Recognition}

\author{
    Cheng-Ze Lu$^{1}$\thanks{Work done when Cheng-Ze Lu and Zhicheng Huang interned at Bytedance Inc.} \ \
    \enskip Xiaojie Jin$^{2\,}$\thanks{Correspondense to: Xiaojie Jin$<$\url{jinxiaojie@bytedance.com}$>$.} \ \
    \enskip Zhicheng Huang$^{3\,*}$ 
    \enskip Qibin Hou$^{1}$ \\
    \enskip \textbf{Ming-Ming Cheng}$^{1}$ 
    \enskip \textbf{Jiashi Feng}$^{2\,}$ \\
$^1$ Nankai University \quad $^2$ 
Bytedance Inc. \quad $^3$ University of Science and Technology Beijing}

\begin{document}
\maketitle

\begin{abstract}
Contrastive Masked Autoencoder (CMAE)~\cite{huang2022contrastive}, as a new self-supervised framework, has shown its potential of learning expressive feature representations in visual image recognition.
This work shows that CMAE also trivially generalizes well on video action recognition without modifying the architecture and the loss criterion.
By directly replacing the original pixel shift with the temporal shift, our CMAE for visual action recognition, CMAE-V for short, can generate stronger feature representations than its counterpart based on pure masked autoencoders.
Notably, CMAE-V, with a hybrid architecture, can achieve 82.2\% and 71.6\% top-1 accuracy on the Kinetics-400 and Something-something V2 datasets, respectively.
We hope this report could provide some informative inspiration for future works.

\end{abstract}

\section{Introduction}
Recent Masked Image Modeling (MIM) methods~\cite{he2022masked,bao2021beit,gao2022convmae}, in the ``mask-and-predict'' style with Vision Transformer~\cite{dosovitskiy2020image}, 
are simple yet capable of achieving promising performance in various downstream tasks.
As for video representation learning, BEVT~\cite{wang2022bevt} is the first work that directly applies MIM for video action recognition and achieves impressive results.
Later, VideoMAE~\cite{tong2022videomae}, ConvMAE~\cite{gao2022convmae}, and OmniMAE~\cite{girdhar2022omnimae} et al. further improve the results.
Despite the strong results, the aforementioned works only focus on learning the relations within each input sample for fulfilling the reconstruction task while neglecting the relations among different samples~\cite{li2022architecture}.

In this report, we show that CMAE~\cite{huang2022contrastive}, which leverages the advantages of both contrastive learning and mask modeling, is also a strong video representation learner without modifying the overall training pipeline or the loss criterion.
To make CMAE adapt to video action recognition, we propose changing the augmentation method to generate positive views to better exploit temporal correlations.
This simple change results in 82.2\% and 71.6\% top-1 accuracy on Kinetics-400 and Something-something V2 datasets.

\begin{figure}[t]
    \centering
    \includegraphics[width=0.98\textwidth]{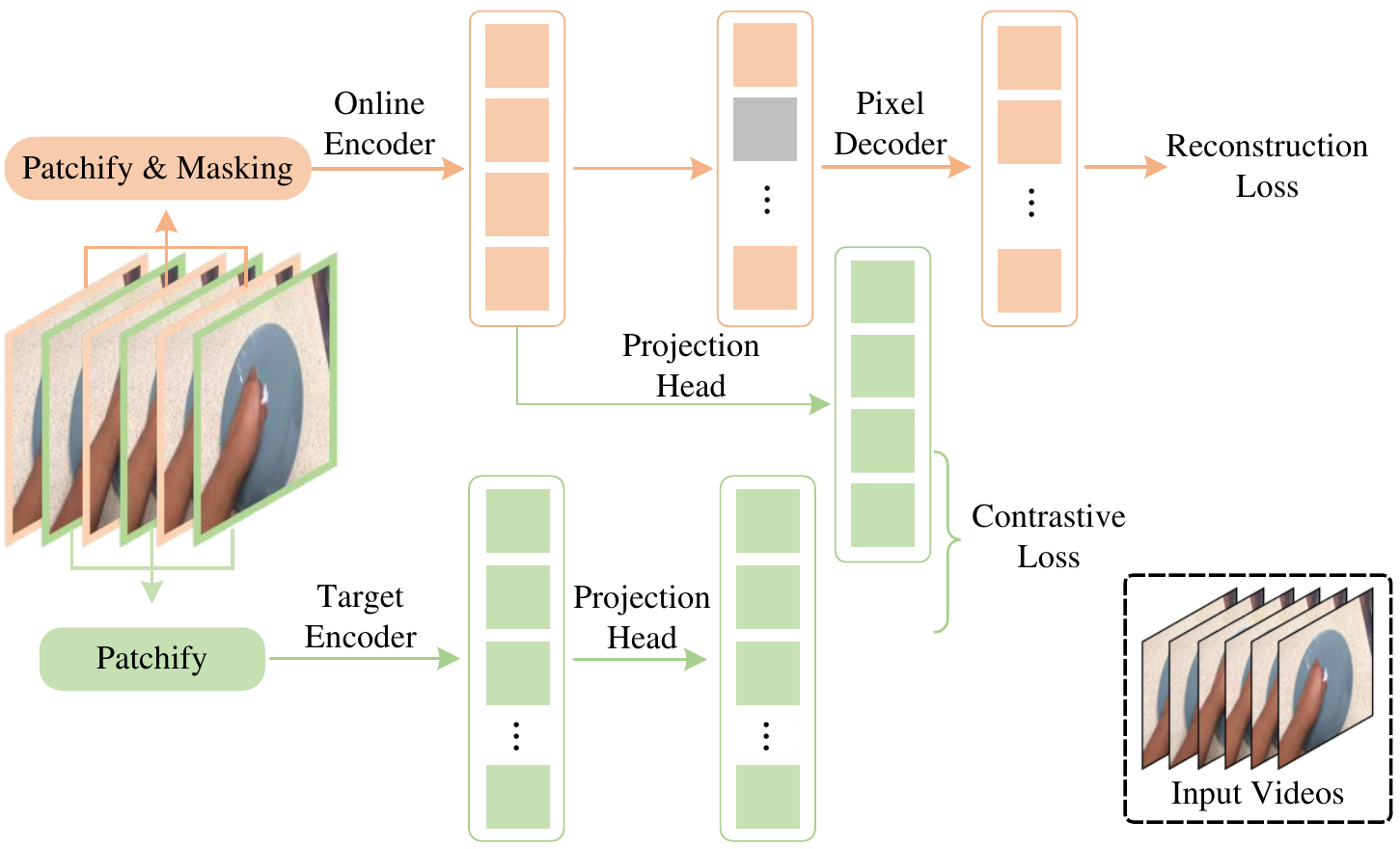}
    \caption{An overview of CMAE-V~\cite{huang2022contrastive} for video representation learning.}
    \label{fig:overview}
\end{figure}

\section{Background}

CMAE~\cite{huang2022contrastive} introduces a contrastive MAE framework for self-supervised representation learning. It adopts a siamese architecture.
One branch is an online updated asymmetric encoder-decoder structure that learns latent representations to reconstruct masked images from a few visible patches, similar to MAE. 
The other one is a momentum encoder that provides contrastive learning supervision.
We now give a quick review of CMAE from the perspective of the training objectives.

\paragraph{Contrastive loss.} 
We assume that the input image $I_s$ to the online encoder $\mathcal F_s$ 
has been tokenized into a token sequence $\{x^{s}_i\}_{i=1}^N$ with $N$ image patches (tokens). 
For its masked version, we denote the visible tokens as $\{x^v_s\}$. 
The embedding features $\{z_{s}^{v}\}$ of visible tokens $\{x^v_s\}$ are obtained by
\begin{equation}
    z_s^v = \mathcal F_s(x_s^v+p_s^v),
\end{equation}
where $p_s^v$ stands for positional embeddings and $F_s$ is the online encoder, which is normally a ViT network.
To align with the output of the target encoder, CMAE uses the feature decoder $\mathcal G_f$ to recover the features of the masked tokens. 
It can be represented as follows:
\begin{equation}
    y_s = \frac{1}{N}\sum \mathcal G_f(z_s^v,z_s^m),
\end{equation}
where $\{z_s^m\}$ are learnable MASK tokens and $y_s$ is the mean-pooled features for calculating the contrastive loss. For succinctness, we omit the index of tokens in above equation.

Similarly, the input tokens to the target encoder $\mathcal F_t$ are denoted as $\{x^t_j\}_{j=1}^N$.
CMAE adopts the mean-pooled features $z_t$ of target encoder's output for simplicity, i.e., 
\begin{equation}
    z_t = \frac{1}{N} \sum_{j=1}^N \mathcal F_t(x^t_j).
\end{equation}

After the frequently used ``projection-prediction'' head~\cite{grill2020bootstrap}, we can get the transformed output $y_{s}^p$ and $z_{t}^p$, respectively.
CMAE first computes the cosine similarity $\rho$ as:
\begin{equation}
    \rho=\frac{y_{s}^p\cdot z_{t}^p}{\left \|y_{s}^p \right \|_{2} \left\|z_{t}^p \right \|_{2}}.
\end{equation}
For calculating the contrastive loss, 
CMAE constructs the positive pairs from the same image and the negative pairs from the different images in a batch.
Then the loss function of InfoNCE loss $L_c$ is calculated as follows:
\begin{equation}
    L_c = -\log \frac{\exp(\rho^{+}/\tau)}{\exp(\rho^{+}/\tau)+\sum_{j=1}^{K-1}(\exp(\rho_j^{-} /\tau))},
\label{eq:infonce}
\end{equation}
where $\tau$ is the temperature constant and $K$ is the batch size.
$\rho^+$ and $\rho^-$ represent the similarity of positive and negative pairs, respectively.

\paragraph{Reconstruction loss.} 
Following~\cite{he2022masked}, CMAE also introduces the pixel decoder $\mathcal G_p$ for 
mapping the latent features $z_s^v$ and MASK tokes $z_s^m$ (shared in contrastive loss) to the
feature space of the target encoder and the original images, i.e.
\begin{equation}
    y^{\prime}_m= \mathbb I \cdot \mathcal G_p(z_s^v, z_s^m),
\end{equation}
where $\mathbb I$ is an indicator to only select the prediction corresponding to masked tokens, and $y^{\prime}_m$ is the output prediction for the masked patches.
Supposing the masked patches of the original image as $y_m$, we can get the reconstruction loss as follows:
\begin{equation}
    L_r = \frac{1}{N_m}\sum(y^{\prime}_m - y_m)^2 ,
\end{equation}
where $N_m$ is the number of masked patches in an image.

The overall learning target of CMAE is a weighted combination of reconstruction loss $L_r$ and contrastive loss $L_c$:
\begin{equation}
    L = L_r + \lambda_c L_c.
\label{eq:total_loss}
\end{equation}

\begin{table*}[t!]
\centering
\tablestyle{4.0pt}{1.1}
\begin{tabular}{lcccccc}
\shline
\textbf{Method} & \textbf{Backbone} & \textbf{Pre-train data} & \textbf{Frames} & \textbf{Views}  & \textbf{Param} & \textbf{Top-1}  \\
\shline
\multicolumn{1}{l}{\textsl{Supervised pre-training} } \\ \hdashline
NL I3D$_{\text{CVPR'18}}$~\cite{wang2018nonlocal} & ResNet101 & IN-1K     & 128 & 10$\times$3  & 62  & 77.3   \\ 
TAM$_{\text{ICCV'21}}$~\cite{liu2021tam} &  ResNet152 &IN-1K  &16 & 4$\times$3 & 59 & 79.3 \\
TDN$_{\text{ICCV'21}}$~\cite{wang2021tdn} & ResNet101$_{\times 2}$ &IN-1K    & 8+16 & 10$\times$3 & 88 & 79.4  \\
Video Swin$_{\text{Arxiv'21}}$~\cite{liu2021videoswin}& Swin-B &IN-1K & 32 & 4$\times$3 & 88 & 80.6   \\
TimeSformer$_{\text{ICML'21}}$~\cite{gberta_2021_ICML} & ViT-B &IN-21K   & 8 & 1$\times$3 & 121 & 78.3  \\ 
TimeSformer$_{\text{ICML'21}}$~\cite{gberta_2021_ICML} & ViT-L &IN-21K     & 96 & 1$\times$3 & 430 & 80.7 \\
ViViT FE$_{\text{ICCV'21}}$~\cite{arnab2021vivit} & ViT-L &IN21K     & 128 & 1$\times$3 & N/A & 81.7  \\
Motionformer$_{\text{NeurIPS'21}}$~\cite{patrick2021motionformer} & ViT-B &IN-21K       & 16 & 10$\times$3 & 109  & 79.7  \\
Motionformer$_{\text{NeurIPS'21}}$~\cite{patrick2021motionformer} & ViT-L &IN-21K       & 32 & 10$\times$3 & 382  & 80.2  \\
ip-CSN$_{\text{ICCV'19}}$~\cite{tran2019csn} & ResNet152 &K400     & 32 & 10$\times$3 & 33 & 77.8  \\ 
SlowFast$_{\text{ICCV'19}}$~\cite{feichtenhofer2019slowfast} &  R101+NL &K400      & 16+64 & 10$\times$3 & 60 & 79.8  \\
MViTv1$_{\text{ICCV'21}}$~\cite{fan2021mvit} & MViTv1-B &K400     & 32 & 5$\times$1 & 37  & 80.2  \\
\hline
\multicolumn{1}{l}{\textsl{Self-supervised pre-training} } \\ \hdashline
VIMPAC$_{\text{Arxiv'21}}$~\cite{tan2021vimpac} & ViT-L & HowTo100M+DALLE    & 10 & 10$\times$3 & 307 & 77.4  \\
BEVT$_{\text{CVPR'22}}$~\cite{wang2022bevt}  & Swin-B & IN-1K+K400+DALLE    & 32 & 4$\times$3 & 88 & 80.6 \\
VideoMAE$_{\text{NeurIPS'22}}$~\cite{tong2022videomae} & ViT-B & K400 & 16 & 5$\times$3 & 87  & 80.7  \\
ConvMAE$_{\text{NeurIPS'22}}$~\cite{gao2022convmae} & ConvViT-B & K400 & 16 & 5$\times$3 & 86  & 81.7  \\
OmniMAE$_{\text{Arxiv'22}}$~\cite{girdhar2022omnimae} & ViT-B & IN-1K+K400 & 16 & 5$\times$3 & 87  & 80.6  \\
ST-MAE$_{\text{NeurIPS'22}}$\cite{feichtenhofer2022masked} & ViT-B & K400 & 16 & 7$\times$3 & 87  & 81.3  \\
\hline
\textbf{CMAE-V (800 epoch)} & ViT-B & K400 & 16 & 5$\times$3 & 87 & \textbf{80.2}  \\
\textbf{CMAE-V (800 epoch)} & ConvViT-B & K400 & 16 & 5$\times$3 & 85 & \textbf{81.6}  \\
\textbf{CMAE-V (1600 epoch)} & ViT-B & K400 & 16 & 5$\times$3 & 87 & \textbf{80.9}  \\
\textbf{CMAE-V (1600 epoch)} & ConvViT-B & K400 & 16 & 5$\times$3 & 85 & \textbf{82.2}  \\
\shline
\end{tabular}
\vspace{-0.5mm}
\caption{Comparison between CMAE-V~\cite{huang2022contrastive} and previous supervised and self-supervised methods on K400 dataset.}
\label{tab:k400}
\end{table*}

\section{CMAE-V}
The overall framework of CMAE~\cite{huang2022contrastive} for videos is illustrated in Figure~\ref{fig:overview}. 
Without modifying the framework or the training objective of CMAE, 
we only adapt the way of augmentation and remove the feature decoder in CMAE for video understanding tasks.

The ``pixel shift'' which is proposed in CMAE generates two correlated augmentation views from the same image by slightly shifting in the spatial dimension. These two views constitute to the inputs of online / target encoders. In CMAE-V, we take into account the temporal dimension in the video domain, and
propose \textbf{temporal shift}, a weakly data augmentation method, for generating the inputs of online/target encoders.

We suppose that for the online branch, 
the input video clip
$\mathbf{V_s}\in \mathbb{R}^{T \times 3 \times H \times W}$ 
sampled from the origin video are with timestamps of
$\{t_1, t_1 + r, t_1 + 2 * r, ..., t_1 + (T - 1) * r \}$,
where $r\in Z^+$ represents the sampling rate.
A disturbance factor $\delta_t$ is randomly sampled from $[0,p], p\in Z^+$,
and then we can get the input of the target encoder
$\mathbf{V_t}\in \mathbb{R}^{T \times 3 \times H \times W}$ with timestamps of
$\{t_1 + \delta_t, t_1 + r + \delta_t, t_1 + 2 * r + \delta_t, ..., t_1 + (T - 1) * r + \delta_t \}$.
Forcing these two views to be moderately correlated helps the model learn temporal invariance and semantically meaningful representations in the contrastive learning objective.  
Afterward, masking and color augmentation are still applied for $V_s$ and $V_t$ respectively. Note that we do not apply pixel shift in the spatial dimension as in CMAE. We find doing so reduces the computation cost in data processing while still retaining the final performance.

\section{Experiments}
Following most previous works~\cite{tong2022videomae,girdhar2022omnimae}, we evaluate our CMAE-V on the Kinetics-400 (K400)~\cite{kay2017kinetics} and Something-something V2 (SSV2)~\cite{goyal2017something} datasets.
The results are respectively shown in Table~\ref{tab:k400} and Table~\ref{tab:ssv2}.

\begin{table*}[t]
\centering
\tablestyle{4.0pt}{1.1}
\begin{tabular}{lcccccc}
\shline
\textbf{Method} & \textbf{Backbone} & \textbf{Pre-train data} & \textbf{Frames} & \textbf{Views}  & \textbf{Param} & \textbf{Top-1}\\
\shline
\multicolumn{1}{l}{\textsl{Supervised pre-training} } \\ \hdashline
TSM$_{\text{ICCV'19}}$~\cite{lin2019tsm}  & \footnotesize{ResNet50$_{\times 2}$}  & {IN-1K}   & 16+16 & 2$\times$3 & 49 & 66.0\\  
TAM$_{\text{ICCV'21}}$~\cite{liu2021tam} &  \footnotesize{ResNet50$_{\times 2}$} &{IN-1K}    & 8+16 & 2$\times$3 & 51 & 66.0 \\
TDN$_{\text{ICCV'21}}$~\cite{wang2021tdn} & \footnotesize{ResNet101$_{\times 2}$} & {IN-1K}     & 8+16 & 1$\times$3 & 88 & 69.6   \\

SlowFast$_{\text{ICCV'19}}$~\cite{feichtenhofer2019slowfast} &  ResNet101 & {Kinetics-400} & 8+32 & 1$\times$3 & 53 & 63.1\\ 
MViTv1$_{\text{ICCV'21}}$~\cite{fan2021mvit} & MViTv1-B &{Kinetics-400} & 64 & 1$\times$3 & 37 & 67.7 \\
TimeSformer$_{\text{ICML'21}}$~\cite{gberta_2021_ICML} & ViT-B & {IN-21K}  & 8 & 1$\times$3 & 121 & 59.5  \\ 
TimeSformer$_{\text{ICML'21}}$~\cite{gberta_2021_ICML} & ViT-L &{IN-21K} & 64  & 1$\times$3 & 430 & 62.4 \\
ViViT FE~\cite{arnab2021vivit} & ViT-L &  IN-21K+K400  & 32 & 4$\times$3 & N/A & 65.9 \\ 
Motionformer$_{\text{NeurIPS'21}}$~\cite{patrick2021motionformer} & ViT-B &  IN-21K+K400    & 16 & 1$\times$3 & 109  & 66.5  \\
Motionformer$_{\text{NeurIPS'21}}$~\cite{patrick2021motionformer} & ViT-L & IN-21K+K400   & 32 & 1$\times$3 & 382  & 68.1  \\
Video Swin$_{\text{Arxiv'21}}$~\cite{liu2021videoswin}  & Swin-B & IN-21K+K400  & 32 & 1$\times$3 & 88 & 69.6\\
\hline
\multicolumn{1}{l}{\textsl{Self-supervised pre-training} } \\ \hdashline

VIMPAC$_{\text{Arxiv'21}}$~\cite{tan2021vimpac} & ViT-L & HowTo100M &10 & 10$\times$3 & 307 & 68.1 \\
BEVT$_{\text{CVPR'22}}$~\cite{wang2022bevt}  & Swin-B & \scriptsize{IN-1K+K400+DALL-E}  & 32 & 1$\times$3 & 88 & 70.6 \\ 
OmniMAE$_{\text{Arxiv'22}}$\cite{girdhar2022omnimae} & ViT-B & IN-1K+SSv2 &16  & $2\times$3 & 87 & 69.5 \\
VideoMAE$_{\text{NeurIPS'22}}$~\cite{tong2022videomae} & ViT-B & SSv2 & 16 & 2$\times$3 & 87 & 70.3 \\
ConvMAE$_{\text{NeurIPS'22}}$~\cite{gao2022convmae} & ConvViT-B & SSv2 & 16 & 2$\times$3 & 86  & 69.9  \\
\hline

\textbf{CMAE-V (800 epoch)} & ViT-B & SSv2 & 16 & 2$\times$3 & 87 & \textbf{69.7}  \\
\textbf{CMAE-V (800 epoch)} & ConvViT-B & SSv2 & 16 & 2$\times$3 & 85 & \textbf{71.1}  \\
\textbf{CMAE-V (1600 epoch)} & ViT-B & SSv2 & 16 & 2$\times$3 & 87 & \textbf{70.5} \\
\textbf{CMAE-V (1600 epoch)} & ConvViT-B & SSv2 & 16 & 2$\times$3 & 85 & \textbf{71.6} \\
\shline
\end{tabular}
\caption{Comparison between CMAE-V~\cite{huang2022contrastive} and previous supervised and self-supervised methods on SSV2 dataset.}
\label{tab:ssv2}

\end{table*}

\subsection{Implementation Details}
We closely follow the settings of VideoMAE~\cite{tong2022videomae} to pre-train and finetune our model.
We use the ViT-B~\cite{dosovitskiy2020image} model as our encoder. 
In addition, we follow ConvMAE~\cite{gao2022convmae}, which replaces the ViT with a hybrid convolutional ViT, to further validate the transferability of CMAE-V to other network structures.
All pre-training and finetuning experiments are conducted on 64 NVIDIA A100 GPUs.

\paragraph{Pre-training.}
During pre-training, a masking ratio of 90\% is adopted following~\cite{tong2022videomae} due to the large temporal redundancy.
We use the AdamW~\cite{loshchilov2017decoupled} optimizer with a batch size of 2048, and the momentum is set to $\beta_1 = 0.9$, $\beta_2 = 0.95$.
The base learning rate is set to 1.5e-4, 
and the linear scaling rule~\cite{goyal2017accurate}: $lr = base\_lr \times batch\_size / 256$ is used.
Cosine learning rate schedule~\cite{loshchilov2016sgdr} with a warmup of 40 epochs is adopted. 

\paragraph{Finetuning.}
During finetuning, we adopt the repeated sampling strategy~\cite{hoffer2020augment} to alleviate long data loading time.
The AdamW~\cite{loshchilov2017decoupled} optimizer is adopted, and the momentum is set to $\beta_1 = 0.9$, $\beta_2 = 0.999$.
Besides, the weight decay is set as 0.05.
We finetune the model for 75/50 epochs for K400/SSV2 with 5 warmup epochs and cosine learning rate schedule~\cite{loshchilov2016sgdr}.

\subsection{Results}
In Table~\ref{tab:k400} and Table~\ref{tab:ssv2}, we compare CMAE with both supervised and self-supervised methods.
CMAE achieves a top-1 accuracy of 80.9\% on K400 dataset, which is 0.2\% higher than VideoMAE~\cite{tong2022videomae}.
On SSV2 dataset, CMAE also slightly surpasses VideoMAE by 0.2\%.
When replacing the vanilla ViT encoder with a hybrid convolutional ViT~\cite{gao2022convmae}, 
CMAE achieves new state-of-the-art on both benchmarks, significantly outperforming ConvMAE by 0.5\% and 1.7\% on K400 and SSV2 respectively.

\bibliographystyle{abbrvnat}
\bibliography{videobib}
\appendix

\end{document}